\documentclass[10pt,journal,compsoc]{IEEEtran}
\usepackage{pifont}

\usepackage{ifpdf}

\ifCLASSOPTIONcompsoc
\usepackage[nocompress]{cite}
\else
\usepackage{cite}
\fi
\ifCLASSINFOpdf
\usepackage[pdftex]{graphicx}
\else
\usepackage[dvips]{graphicx}
\fi
\usepackage{amsmath}
\usepackage{algorithmic}

\usepackage{array}

\ifCLASSOPTIONcompsoc
\usepackage[caption=false,font=footnotesize,labelfont=sf,textfont=sf]{subfig}
\else
\usepackage[caption=false,font=footnotesize]{subfig}
\fi
\usepackage{url}

\usepackage{caption}
\usepackage{color}
\usepackage{lipsum}
\usepackage{algorithm}
\usepackage{tikz}
\usepackage{comment}
\usepackage{amsmath,amssymb} 
\usepackage{color}
\usepackage{algorithm}
\usepackage{multirow}
\usepackage{multicol}
\usepackage{listings}
\usepackage{etoolbox}
\usepackage{color,colortbl}
\definecolor{lightgray}{HTML}{F0F0F0}  
\definecolor{rowbackground}{HTML}{F9F9F9}
\definecolor{LightCyan}{rgb}{0.88,1,1}
\definecolor{myblue}{rgb}{.935,.935,.99}

\usepackage[accsupp]{axessibility}  

\usepackage{epsfig}
\usepackage{graphicx}
\usepackage{amsmath}
\allowdisplaybreaks[4]
\usepackage{amssymb}   	
\usepackage{colortbl}
\usepackage{url}
\usepackage[utf8]{inputenc} 
\usepackage[T1]{fontenc}    
\usepackage{booktabs}       
\usepackage{amsfonts}       
\usepackage{nicefrac}       
\usepackage{microtype}      
\usepackage{xcolor}
\usepackage{bm}
\usepackage{bbm}
\usepackage{amsbsy}
\usepackage{amsmath}
\usepackage{cases}
\usepackage{enumitem}
\usepackage{lipsum}
\usepackage{graphicx}
\usepackage{multirow}
\usepackage{grffile}
\usepackage{diagbox}
\usepackage{pifont}
\usepackage{amssymb}
\usepackage{multirow}
\usepackage{multicol}
\newcolumntype{I}{!{\vrule width 1pt}}
\usepackage{color,colortbl}
\definecolor{lightgray}{HTML}{F0F0F0}  
\definecolor{rowbackground}{HTML}{F9F9F9}
\definecolor{Gray}{gray}{0.9}
\definecolor{LightCyan}{rgb}{0.88,1,1}
\definecolor{myblue}{rgb}{.935,.935,.99}
\usepackage{textpos}
\usepackage{cite}
\newcolumntype{x}[1]{>{\centering\arraybackslash}p{#1pt}}

\usepackage{epsfig}
\usepackage{graphicx}
\usepackage{amsmath}
\usepackage{amssymb}

\usepackage{booktabs}
\usepackage{tabulary}
\usepackage{soul}
\usepackage{xspace}\xspace
\usepackage{adjustbox}
\usepackage{array}

\usepackage{dblfloatfix}
\usepackage{transparent}

\usepackage{algorithm}
\usepackage{listings}
\usepackage{ragged2e}

\makeatletter
\DeclareRobustCommand\onedot{\futurelet\@let@token\@onedot}
\def\@onedot{\ifx\@let@token.\else.\null\fi\xspace}

\def\eg{\emph{e.g}\onedot} 
\def\ie{\emph{i.e}\onedot}

\def\etal{\emph{et al}\onedot}
\makeatother

\makeatletter
\AfterEndEnvironment{algorithm}{\let\@algcomment\relax}
\AtEndEnvironment{algorithm}{\kern2pt\hrule\relax\vskip3pt\@algcomment}
\let\@algcomment\relax
\newcommand\algcomment[1]{\def\@algcomment{\footnotesize#1}}
\renewcommand\fs@ruled{\def\@fs@cfont{\bfseries}\let\@fs@capt\floatc@ruled
	\def\@fs@pre{\hrule height.8pt depth0pt \kern2pt}%
	\def\@fs@post{}%
	\def\@fs@mid{\kern2pt\hrule\kern2pt}%
	\let\@fs@iftopcapt\iftrue}
\makeatother

\newcommand{\cgaphl}[2]{
	\fontsize{8pt}{1em}\selectfont{\textcolor{Highlight}{($\textbf{#1}$\textbf{#2})}}
}
\newcommand{\cgaphll}[2]{
	\fontsize{7pt}{1em}\selectfont{\textcolor{citecolor}{($\textbf{#1}$\textbf{#2})}}
}
\newcolumntype{x}[1]{>{\centering\arraybackslash}p{#1pt}}
\newcolumntype{y}[1]{>{\raggedright\arraybackslash}p{#1pt}}
\newcolumntype{z}[1]{>{\raggedleft\arraybackslash}p{#1pt}}
\definecolor{Highlight}{HTML}{39b54a}  
\definecolor{Gray}{gray}{0.92}

\makeatletter
\newcommand*\bigcdot{\mathpalette\bigcdot@{.5}}
\newcommand*\bigcdot@[2]{\mathbin{\vcenter{\hbox{\scalebox{#2}{$\m@th#1\bullet$}}}}}
\makeatother

\newcommand{\ff}{\boldsymbol{f}}
\newcommand{\bx}{\boldsymbol{x}}
\newcommand{\bs}{\boldsymbol{s}}
\newcommand{\bz}{\boldsymbol{z}}

\newcommand{\mP}{\mathcal{P}}

\newcommand{\mQ}{\mathcal{Q}}

\newcommand{\later}[1]{}
\newcommand{\longer}[1]{}	%

\def\BE{\begin{equation}}
\def\EE{\end{equation}}
\def\BEA{\begin{eqnarray}}
\def\EEA{\end{eqnarray}}
\def\BEAS{\begin{eqnarray*}}
\def\EEAS{\end{eqnarray*}}

\definecolor{citecolor}{HTML}{0071bc}

\begin{document}
	\title{Rebalanced Siamese Contrastive Mining \\for Long-Tailed Recognition}
	\author{Zhisheng~Zhong,~\IEEEmembership{Student Member,~IEEE,} Jiequan~Cui,~\IEEEmembership{Student Member,~IEEE,}
		Zeming~Li,~Eric~Lo,~Jian~Sun,~Jiaya~Jia,~\IEEEmembership{Fellow,~IEEE}
		
		\IEEEcompsocitemizethanks{
			
			\IEEEcompsocthanksitem Z.~Zhong J.~Cui, E.~Lo, and J.~Jia are with the Department of Computer Science \& Engineering, The Chinese University of Hong Kong, ShaTin, Hong Kong. E-mail: \{ zszhong21, jqcui, ericlo, leojia\}@cse.cuhk.edu.hk
			
			\IEEEcompsocthanksitem Z. Li and J. Sun are with the MEGVII Technology. Email: \{lizeming,
				sunjian\}@megvii.com}}

	\IEEEtitleabstractindextext{%
		\begin{abstract}
			\justifying Deep neural networks perform poorly on heavily class-imbalanced datasets. Given the promising performance of contrastive learning, we propose \textbf{Re}balanced \textbf{S}iamese \textbf{Co}ntrastive \textbf{M}ining (\textbf{ResCom}) to tackle imbalanced recognition. 
			Based on the mathematical analysis and simulation results, we claim that supervised contrastive learning suffers a dual class-imbalance problem at both the original batch and Siamese batch levels, which is more serious than long-tailed classification learning. 
			In this paper, at the original batch level, we introduce a class-balanced supervised contrastive loss to assign adaptive weights for different classes. At the Siamese batch level, we present a class-balanced queue, which maintains the same number of keys for all classes.  
			Furthermore, we note that the imbalanced contrastive loss gradient with respect to the contrastive logits can be decoupled into the positives and negatives, and easy positives and easy negatives will make the contrastive gradient vanish. We propose supervised hard positive and negative pairs mining to pick up informative pairs for contrastive computation and improve representation learning.
			Finally, to approximately maximize the mutual information between the two views, we propose Siamese Balanced Softmax and joint it with the contrastive loss for one-stage training. Extensive experiments demonstrate that ResCom \textit{outperforms the previous methods by large margins} on multiple long-tailed recognition benchmarks. Our code and models are made publicly available at: \textbf{\url{https://github.com/dvlab-research/ResCom}}.
			
		\end{abstract}
		}
	
	\maketitle
	\IEEEdisplaynontitleabstractindextext
	\IEEEpeerreviewmaketitle

	\IEEEraisesectionheading{\section{Introduction}\label{intro}}

\begin{figure*}[!t]
	\centering
	\includegraphics[width=0.99\linewidth]{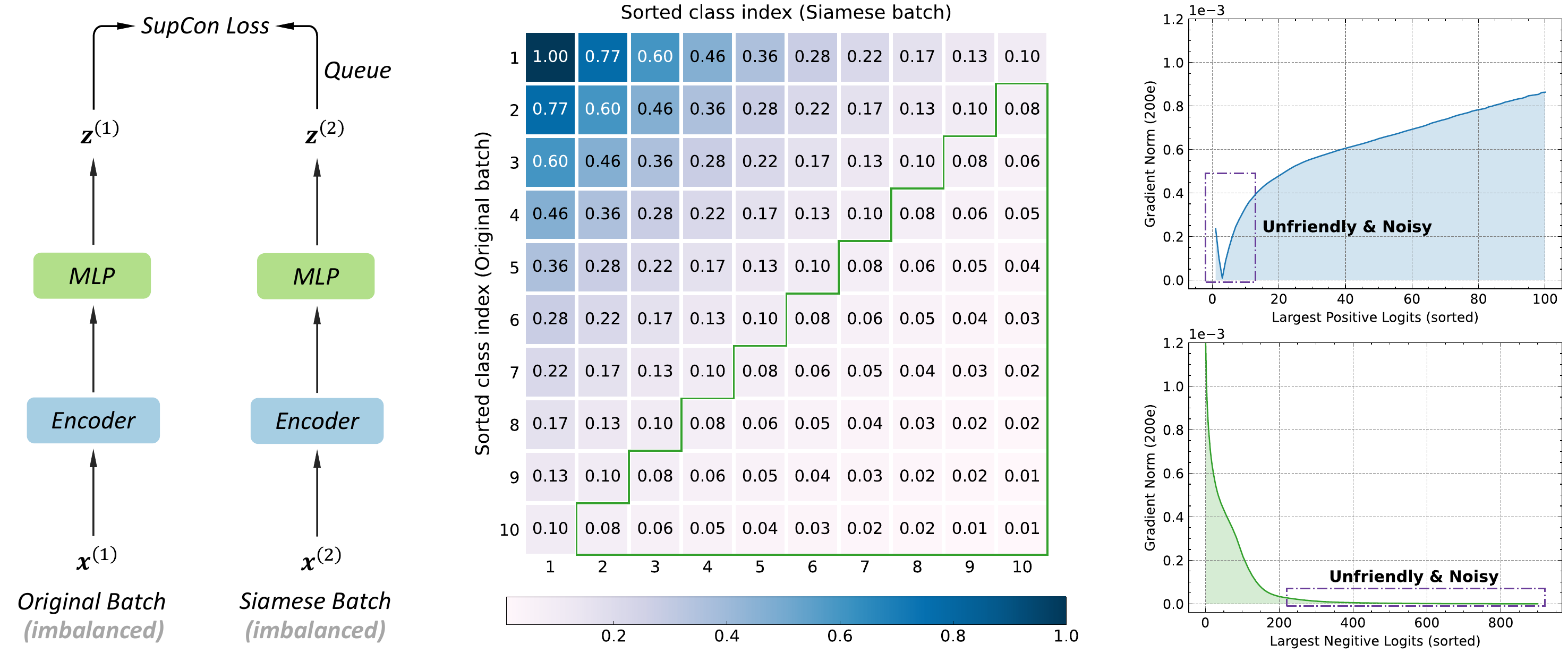}
	\caption{Left: illustration of the original batch, the Siamese batch, and the queue in the contrastive learning framework. \textit{These three parts are all imbalanced} on long-tailed datasets.  Center: frequency visualization of contrastive pairs for all class combinations normalized by the maximal frequency of contrastive pairs, \ie, the total number of pairs from (Class-1, Class-1). The sampling simulation is run for 200 epochs on \textbf{CIFAR-10-LT, imbalanced factor 10}. The classes are sorted by descending values of the number of samples per class. The imbalance issue of supervised contrastive learning comes from \textit{two levels}: the \textit{original batch} and the \textit{Siamese batch (or queue)}. Right: the average gradient norm for the SupCon loss with respect to the positives and negatives (sorted by similarity in descending order). The model is trained on CIFAR-10-LT, imbalanced factor 10 for 200 epochs. The memory queue contains about 100 positives and 900 negatives. \textit{The gradient of most easy positives and negatives could be vanished and may be noisy for representation learning}.}
	\label{fig:intro}
\end{figure*}
	
	\IEEEPARstart{W}ITH the emergence of powerful GPUs, and numerous available large-scale and high-quality datasets such as ImageNet~\cite{imagenet}, COCO~\cite{coco}, and Places~\cite{places}, deep neural networks~(DNNs) have shown great success to many visual discriminative tasks, including image recognition~\cite{resnet}, object detection~\cite{fasterrcnn}, and semantic segmentation~\cite{cordts2016cityscapes}. The above datasets are all carefully constructed and approximately balanced concerning the number of instances for each class or object. However, real-world problems typically exhibit a long-tailed distribution, where a few classes contain plenty of samples but the others are associated with only a few samples. Learning in such a real-world case is challenging as the low-frequency classes can be easily overwhelmed by high-frequency ones, which makes DNNs suffer from significant performance degradation.

	Recently, contrastive learning has shown great promise in unsupervised representation learning~\cite{caron2020unsupervised, simsiam}. Supervised contrastive learning~(SupCon) is an extension to contrastive learning by incorporating the label information to compose positive and negative pairs~\cite{supcon}. SupCon provides consistent boosts in top-1 accuracy compared with conventional cross-entropy loss. Moreover, it is also more robust to natural corruptions. However, to ensure performance, SupCon needs a large batch size for constructing enough contrastive pairs, which is extremely computation and memory costly. When using with memory based alternatives~\cite{moco} (queue), SupCon can achieve better recognition performances and allows to use a small batch size for contrastive training, which significantly reduces compute and memory footprint.
	
	Though supervised contrastive learning works well in a balanced setting, for imbalanced datasets, SupCon suffers a serious \textit{dual class-imbalance issue} at both the \textit{original batch-level} and the \textit{Siamese batch-level} (the left part of Fig.~\ref{fig:intro} for illustration).  Here we provide a brief mathematical explanation (\textit{a rigorous mathematical proof is shown in Sec.~\ref{preliminaries}}): suppose the batch size is $B$, the frequency for Class-$k$ of the training dataset is $\pi_k$. For a total batch, it approximately contains $B\pi_{\text{max}}$ samples from the most frequent class for both the original and the Siamese batch, which constructs $B\pi_{\text{max}}\cdot B\pi_{\text{max}}$ positive pairs for the most frequent class. Similarly, SupCon includes $B\pi_{\text{min}}\cdot B\pi_{\text{min}}$ positive pairs of the least frequent class in a total batch. Thus, the imbalanced factor for the contrastive domain can be approximately computed by $\frac{B\pi_{\text{max}}\cdot B\pi_{\text{max}}}{B\pi_{\text{min}} \cdot B\pi_{\text{min}}}=\bigl(\frac{\pi_{\text{max}}}{\pi_{\text{min}}}\bigr)^2$, where $\frac{\pi_{\text{max}}}{\pi_{\text{min}}}$ is the imbalanced factor for classification learning. It means the degree of imbalance for supervised contrastive learning is \textit{quadratic} to classification learning. Likewise, we can derive the imbalanced factor for the memory queue version of SupCon: suppose the queue size is $Q$ and \textit{queue is enqueued from the Siamese batch},  the imbalanced factor can be written as $\frac{B\pi_{\text{max}}\cdot Q\pi_{\text{max}}}{B\pi_{\text{min}} \cdot Q\pi_{\text{min}}}=\bigl(\frac{\pi_{\text{max}}}{\pi_{\text{min}}}\bigr)^2$, which meets the same conclusion. The simulation experiment shown in Fig.~\ref{fig:intro} (center) also verifies the above proposition: the related frequency of many contrastive pairs (in the green region) can be less than $0.1$ (the reciprocal of classification imbalanced factor), and the minimal related frequency closes to $0.01$. Thus, long-tailed contrastive learning suffers a \textit{more difficult imbalance issue} and we should explore new ways to deal with.
	
	In this work, we present \textbf{Re}balanced \textbf{S}iamese \textbf{Co}ntrastive \textbf{M}ining (\textbf{ResCom}) to tackle the dual contrastive imbalance issue: \textbf{(i)}~We propose a class-balanced contrastive loss, which \textit{assigns adaptive weights for different classes}. Under this case, the network will pay more attention to the tail classes samples to relieve the original batch-level imbalance. \textbf{(ii)}~We maintain the dictionary as a class-balanced queue: \textit{all classes contain the same number of keys}. When computing the SupCon loss, each class will construct the same number of positive and negative contrastive pairs. It can greatly alleviate the Siamese batch-level imbalance. 
	
	Furthermore, we note that the contrastive loss gradient with respect to the contrastive logits can be decoupled into the positive and negative parts. Besides, as shown in Fig.~\ref{fig:intro} (right), \textit{most easy positive and easy negative pairs make the contrastive gradient vanish and damage or disturb the learning process}. We propose \textbf{S}upervised hard positive and negative \textbf{P}airs \textbf{M}ining~(\textbf{SPM}) to pick up these useful and informative positive pairs and negative pairs, and enhance the representation learning. 
	
	Finally, to approximately \textit{maximize the mutual information of features from the two views} on long-tailed datasets, we propose \textbf{Siam}ese \textbf{B}alanced \textbf{S}oftmax~(\textbf{SiamBS}), which can be \textit{effectively and jointly trained with a contrastive loss in one-stage} for better recognition performances. 
	
	Ablation studies show ResCom can \textit{obtain more robust and generalized results} and \textit{improve the performances of the tail and medium classes while maintaining the performance of the head classes} compared with its counterparts. On multiple popular long-tailed recognition benchmark datasets, ResCom can \textit{consistently surpass} the previous methods by \textit{large margins}, demonstrating the effectiveness of ResCom.

	\section{Related Work}
	
	\vspace{0.5em}
	\noindent
	\textbf{Contrastive learning.} Contrastive learning is a framework that learns similar and dissimilar representations from data that are organized into similar and dissimilar pairs, respectively. Recently, contrastive learning has shown great promise in unsupervised representation learning~\cite{oord2018representation, simclr, moco}. Chen \etal~\cite{simclr} proposed SimCLR, and is the first to match the performance of a supervised ResNet~\cite{resnet} with only a linear classifier trained on self-supervised representation on large-scale datasets. He \etal proposed MoCo~\cite{moco, mocov2}, which uses a momentum encoder to maintain consistent representations of negative pairs drawn from a memory bank. Without using negative pairs, Grill \etal proposed BYOL~\cite{byol}, which uses a momentum network to produce prediction targets as a means of stabilizing the bootstrap step. Supervised contrastive learning~\cite{supcon} is an extension to contrastive learning by incorporating the label information to compose positive and negative pairs, which can get better feature representation than cross-entropy.

	\vspace{0.5em}
	\noindent
	\textbf{Resampling and reweighting.} Resampling and reweighting are the most intuitive methods to deal with long-tailed recognition. There are two groups of resampling strategies: over-sampling the tail classes~\cite{shen2016relay, byrd2019effect} and under-sampling the head classes~\cite{japkowicz2002class, buda2018systematic}. Reweighting~\cite{huang2016learning, wang2017learning} is another prominent strategy. It assigns different weights for classes and even instances. The vanilla reweighting method gives class weights in reverse proportion to the number of samples of classes. However, with large-scale data, reweighting makes deep models difficult to optimize during training. Cui \etal~\cite{effnum} relieved the problem by using a effective number to calculate the class weight.  Another line of work~\cite{lin2017focal, tan2020equalization} is to adaptively re-weight each instance.

	\vspace{0.5em}
	\noindent
	\textbf{Loss margin modification.}
	Loss margin modification seeks to handle class imbalance by adjusting the decision margin. Cao \etal proposed LDAM~\cite{ldam} to integrate per-class margin into the cross-entropy loss. Balanced Softmax~\cite{balsfx} proposed to use the label frequencies to adjust model predictions. Following the similar idea, Menon \etal proposed logit adjustment to post-hoc shift the model logits based on label frequencies~\cite{logitsadj}. LADE~\cite{lade}
	proposed to use the label frequencies of test data to post-adjust the model outputs, so that the trained model can be calibrated for arbitrary test distribution.
	
	\vspace{0.5em}
	\noindent
	\textbf{Multi-stage methods.}
	Two-stage methods are shown to be effective for long-tailed recognition recently. Cao \etal~\cite{ldam} proposed deferred reweighting and deferred re-sampling, working better than conventional one-stage methods. Its second stage, starting from better features, adjusts the decision boundary and locally tunes features. Kang \etal~\cite{lws} and Zhou \etal~\cite{bbn} concluded that although class re-balance matters for jointly training representation and classifier, instance-balanced sampling gives more general representations. Thus, Kang \etal~\cite{lws} proposed the two-stage decoupling model for decomposing representation and classifier learning. Zhong \etal~\cite{mislas} proposed shift learning and class-aware label smoothing to further improve the second stage performances.

	\vspace{0.5em}
	\noindent
	\textbf{Long-tailed contrastive framework.}
	Long-tailed contrastive recognition is also explored by many researchers. Yang \etal~\cite{yang2020rethinking} proposed self-supervised pre-training, which is the first to use self-supervised learning (\eg, contrastive learning~\cite{moco} or rotation prediction~\cite{gidaris2018unsupervised}) for model pre-training, followed by standard training on long-tailed data. Kang \etal~\cite{kang2020exploring} proposed a $k$-positive contrastive loss to learn a balanced feature space, which helps to alleviate class imbalance and improve model generalization. Following that, Hybrid network~\cite{wang2021contrastive} introduced a prototypical contrastive learning strategy to enhance long-tailed learning.  DRO-LT~\cite{samuel2021distributional} extended the prototypical contrastive learning with distribution robust optimization, which makes the learned model more robust. PaCo~\cite{paco} further innovated supervised contrastive learning by adding a set of parametric learnable class centers, which play the same role as a classifier if regarding the class centers as the classifier.

	\begin{figure*}[t]
		\centering
		\includegraphics[width=0.99\textwidth]{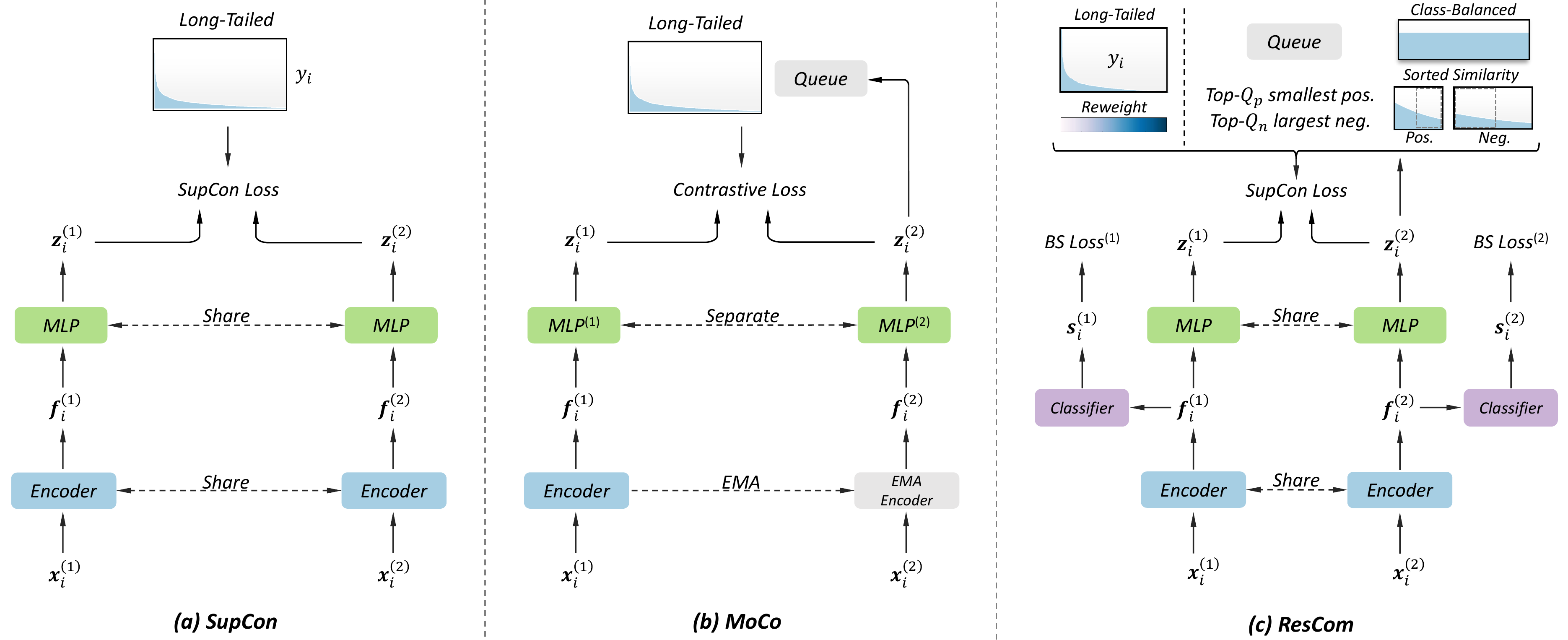} 
		\caption{Conceptual comparison of three contrastive mechanisms for long-tailed recognition (empirical comparisons are listed in Sec.~\ref{sec_ab}). (a) The SupCon encoders for computing the contrastive pairs (based on \textit{the long-tailed distribution labels}) are updated end-to-end by back-propagation (the two encoders are shared). (b) MoCo encodes the new keys by a momentum-updated encoder, and maintains a \textit{long-tailed distributed queue} of keys. (c) ResCom tackles the serious long-tailed contrastive learning issue by a class-balanced contrastive loss, a class-balanced queue, supervised hard positive and negative pairs mining, and Siamese Balanced Softmax for \textit{one-stage end-to-end} training.}
		\label{fig_framework}
	\end{figure*}
	
	\section{Rebalanced Siamese Contrastive Mining}
	
	\subsection{Preliminaries}\label{preliminaries}
	\vspace{0.5em}
	\noindent
	\textbf{Supervised contrastive learning.}
	Khosla \etal \cite{supcon} extended the self-supervised contrastive loss with \textit{label information} into a supervised contrastive loss. Within a multi-viewed batch (batch size is $B$), let $i \in \{1,...,B\}$ be the index of an arbitrary sample. For a training sample $\{\bx_i, y_i\}$, two separate data augmentation operators are sampled from the same family of augmentations and applied to each data example $\bx_i$ to obtain two correlated views $\bx_i^{(1)}$ and $\bx_i^{(2)}$.  $\bx^{(1)}_i$ and $\bx^{(2)}_i$ share the same class label $y_i$. For simplification, we define $y_i = y_i^{(1)}=y_i^{(2)}$ here. In the following, an encoder network, \eg, ResNet~\cite{resnet}, is shared between two Siamese branches to learn the image features, $\ff_i^{(1)}$ and $\ff_i^{(2)}$ of two views. A MLP projection model and the $l_2$ normalization map the image features $\ff_i^{(1)}$ and $\ff_i^{(2)}$ into the embedding representations $\bz_i^{(1)}$ and $\bz_i^{(2)}$ for contrastive learning. Finally, an InfoNCE~\cite{oord2018representation} style loss function is adopted on the embedding representations $\bz$ and labels $y$.

	\vspace{0.5em}
	\noindent
	\textbf{Dynamic memory queue.}
	The supervised contrastive loss usually \textit{requires a large batch size}, \eg, $B = 6,144$, for training to guarantee good performances, which requires a lot of computing resources and memory. He \etal \cite{moco} proposed a dynamic memory queue for mini-batch contrastive learning. \cite{moco} maintained a dictionary as a queue of embedding representation samples ($\bz$) coming from several preceding mini-batches. The queue decouples the dictionary size from the mini-batch size. Here we present the queue vision of the supervised contrastive loss ${L}^{\rm{SupCon}}_i$ as:
	\begin{align}
		\label{eq:sc}
		{L}^{\rm{SupCon}}_i&= -\frac{1}{|\mP_i|}\sum_{\bz_p\in \mP_i}\log\frac{\exp{(\bz^{(1)}_i} \bigcdot \bz_p / \tau)}{\sum\limits_{\bz_q \in \mQ}\exp{(\bz^{(1)}_i} \bigcdot \bz_q / \tau)},\\
		\mQ&=\{\bz^{(2)}_i \text{ from several preceding mini-batches}\},  \notag\\ 
		\mP_i&=\{\bz \in \mQ \land \text{the class label of } \bz = y_i\}, \notag
	\end{align}
	where $|\mP_i|$ is its cardinality, the $\bigcdot$ symbol denotes the inner dot product, and $\tau$ is the temperature.
	
	\vspace{0.5em}
	\noindent
	\textbf{Disadvantages.}
	If we directly apply the queue vision of supervised contrastive loss, Eq.~(\ref{eq:sc}), to long-tailed recognition, it will have the following problems: \textbf{(i)}~\textit{the original batch-level imbalance} and \textbf{(ii)} \textit{the Siamese batch-level (or queue-level) imbalance}. As shown in Fig.~\ref{fig_framework}(a), for the original batch-level imbalance, we consider the contrastive loss of a  total batch: ${L}^{\rm{SupCon}} = \frac{1}{B}\sum_{i=1}^{B}{L}^{\rm{SupCon}}_i$. Each ${L}^{\rm{SupCon}}_i$ just focuses on positive pairs from Class-$y_i$. Due to the imbalanced class distribution, it makes the network pay more attention to the contrastive pairs of the head classes rather than the tailed classes. Additionally, in Fig.~\ref{fig_framework}(b), for the the Siamese batch-level imbalance, the queue is enqueued by the preceding several mini-batches of embedding samples from View-2, \ie, $\bx^{(1)}, \bz^{(1)}$. The queue also contains more embedding samples from the head classes than the tail classes. 
	
	\vspace{0.5em}
	\noindent
	\textbf{Problem definition.}
	Given a dataset $\mathcal{D}$ with totally $N$ samples and $K$ categories, where each category contains $n_1, n_2, \ldots, n_K$ samples, respectively~(w.l.o.g, $N_1 \geq N_2 \geq \ldots \geq N_K$). Let $\mathcal{B}_t = \{ b_{t,i} \}_{i=1}^{B}$ denote the $t$-th batch set obtained by random sampling, where $b_{t,i}$ is an instance in the batch  $\mathcal{B}_t$, $1 \leq t \leq T$,  $T$ represents the total number of batches in one epoch, and $B$ is the batch size. We also define $B_{t,k}$ as the number of samples which belongs to the Class-$k$ in batch $\mathcal{B}_t$. For memory queue,  $\mathcal{Q}_t = \{ q_{t,i} \}_{i=1}^{Q}$ denote the memory bank composed of $Q$ samples,  where $b_{t,i}$ is an instance in the queue $\mathcal{Q}_t$. We also define $Q_{t,k}$ as the number of samples which belongs to the Class-$k$ in $\mathcal{Q}_t$. Thus, $P_{t,k}=B_{t,k}Q_{t,k}$ denotes the number of sample pairs $\left( b_{t,i}, q_{t,j} \right)$ in which $b_{t,i}$, $q_{t,j}$ belong to the same Class-$k$ (a positive pair). The contrastive imbalance factor $\gamma$ can be defined as:
	\begin{equation}
			\gamma = {\max\limits_{k} \left[\mathbb{E}_{Q_{t,k}} \left( \sum\limits_{t=1}^T  P_{t,k}\right)\right]} \bigg/ {\min\limits_{k} \left[ \mathbb{E}_{Q_{t,k}} \left( \sum\limits_{t=1}^T  P_{t,k}\right) \right]}.  \notag
	\end{equation}

	\vspace{0.5em}
	\noindent
	\textbf{Proposition.}
	Assume that each batch $\mathcal{B}_t$ in the training process is randomly sampled from the dataset $\mathcal{D}$ (without replacement), and the memory bank $\mathcal{Q}_t$ is enqueued by the previous batches. Thus, the random variable $Q_{t,k}$ follows a hypergeometric distribution $H(N, Q, N_k)$, and its expected value is $\frac{N_kQ}{N}$. Then, we can get:
	\begin{equation}
		\mathbb{E}_{Q_{t,k}}\left(\sum_{t=1}^T P_{t,k}\right) =\sum_{t=1}^TB_{t,k}\mathbb{E}\left(  Q_{t,k} \right)=N_k\mathbb{E}\left( Q_{t,k}\right)=\frac{N_k^2Q}{N}. \notag
	\end{equation}
	
	Thus, according to the definition, the contrastive imbalance factor $\gamma$ of the dataset (in one epoch) is: 
	\begin{equation}
		\gamma = \frac{\max\limits_{k} \left[\mathbb{E}_{Q_{t,k}} \left( \sum\limits_{t=1}^T  P_{t,k}\right)\right]}{\min\limits_{k} \left[ \mathbb{E}_{Q_{t,k}} \left( \sum\limits_{t=1}^T  P_{t,k}\right) \right]} = \frac{\max_k (\frac{N_k^2Q}{N})}{\min_k (\frac{N_k^2Q}{N})}= \frac{N_1^2}{N_K^2}=\frac{\pi^2_{1}}{\pi^2_K}.  \notag
	\end{equation}

	
	Overall, both the original batch-level and Siamese batch-level imbalance will lead long-tailed contrastive learning more difficult. As analyzed in Sec.~\ref{intro}, the imbalanced factor for classification is $\frac{\pi_{\text{max}}}{\pi_{\text{min}}}$, while the imbalanced factor for contrastive learning becomes $(\frac{\pi_{\text{max}}}{\pi_{\text{min}}})^{2}$. To solve the above issues, we propose rebalanced Siamese contrastive mining.

	\subsection{Original Batch-Level: Class-Balanced SupCon}
	To solve the batch-level imbalance issue, the most intuitive method is to directly use label frequencies of training samples for loss reweighting. Class-balanced cross-entropy loss~\cite{effnum} introduced the novel concept of \textit{effective numbers} to approximate the expected sample number of different classes. The effective number is an exponential function of the training sample number. Following this concept, we also introduce an effective number to reweight the supervised contrastive loss:
	\begin{equation}
		\label{eqn:cbsc}
		\begin{split}
			{L}^{\rm{CB}}_i &= -\frac{w_{y_i}}{|\mP_i|}\sum_{\bz_p\in \mP_i}\log\frac{\exp{(\bz^{(1)}_i} \bigcdot \bz_p / \tau)}{\sum\limits_{\bz_q \in \mQ}\exp{(\bz^{(1)}_i} \bigcdot \bz_q / \tau)}, \\
			w_{y_i} & = \frac{1 - \beta}{1 - \beta^{N_{y_i}}},
		\end{split}
	\end{equation}
	\noindent where $\beta \in [0, 1)$ is a hyper-parameter. $\frac{1 - \beta^{N_{y_i}}}{1 - \beta}$ is the effective number of samples for Class-$y_i$. Under the class-balanced contrastive loss, tailed class samples will be \textit{assigned larger weights} to relieve the original batch-level imbalance issue.

	\subsection{Siamese Batch-Level: Class-Balanced Queue}\label{bq}
	To solve the Siamese batch-level imbalance issue, we propose a class-balanced queue for imbalanced contrastive learning. We suppose the number of classes is $K$.  In our class-balanced queue model, we maintain $K$ queues for each class separately. For each $\mQ_k$, it contains $Q$ contrastive logit samples only from the $k$-th class, \ie,  $|\mQ_k|=Q, \forall  k=1,2,...,K, \text{ the class label of } \bz \text{ equals } k, \forall \bz \in \mQ_{k}$. The contribution of each category to the class-balanced queue is the same. Thus, the total size of contrastive pairs equals $KQ$, and the balanced queue version of contrastive loss becomes:
	\begin{equation}
		{L}^{\rm{BQ}}_i = -\frac{w_{y_i}}{Q}\sum_{\bz_p \in \mQ_{y_i}}\log\frac{\exp{(\bz^{(1)}_i} \bigcdot \bz_p / \tau)}{\sum\limits_{k=1}^{K}\sum\limits_{\bz_q \in \mQ_k}\exp(\bz^{(1)}_i \bigcdot \bz_q / \tau)}.
	\end{equation}
	\noindent By adopting a class-balanced queue into contrastive learning, the Siamese batch-level imbalance issue can be solved very well. Regardless of the head classes or the tail classes sample, when calculating the contrastive loss of a query, it will have \textit{the same number of positive pairs and negative pairs}.

	\subsection{Supervised Hard Positive and Negative Pairs Mining} \label{sec:spm}
	Here we analyze the class-balanced queue version of contrastive loss from the gradient view. The gradient for ${L}^{\rm{BQ}}_i$ with respect to the embedding $\bz_i^{(1)}$ has the following form:

	\begin{equation}
			\frac{\partial {L}^{\rm{BQ}}_i }{\partial \bz^{(1)}_i} = \frac{w_{y_i}}{\tau}\Biggl[\ \underbrace{\sum_{\bz_p \in \mQ_{y_i}}\bz_p\left(P_{ip}\!-\!\frac{1}{Q}\right)}_{\text{positive pairs}}\!+\!\underbrace{\sum_{k\neq y_i}\sum_{\bz_n \in \mQ_k }\bz_nP_{in}}_{\text{negative pairs}}\ \Biggl], \notag
	\end{equation}

	\begin{equation}
		\label{eq_grad}
		P_{ij}=\frac{\exp{(\bz^{(1)}_i} \bigcdot \bz_j / \tau)}{\sum\limits_{k=1}^{K}\sum\limits_{\bz_q \in \mQ_k}\exp(\bz^{(1)}_i \bigcdot \bz_q / \tau)},
	\end{equation}
	
	\noindent From Eq.~(\ref{eq_grad}), the total gradient can be \textit{decoupled into the positive pairs part and the negative pairs part}. Many studies~\cite{henaff2020data, tian2020contrastive, moco, simclr} show that the number of positive pairs and negative pairs for each batch seriously affects the final recognition accuracy and show increased performance with an appropriate number of positives and negatives, wherein the ability to discriminate between signal and noise (negatives) is improved. On the other hand, \cite{chuang2020debiased, robinson2020contrastive, kalantidis2020hard} argue for the value of hard negatives in \textit{unsupervised contrastive representation learning}: by mining harder negatives, one can get higher performance after training for fewer epochs. However, in SupCon, as shown in Fig.~\ref{fig:intro} (right), \textit{for most easy positive pairs, $P_{ip}\to \frac{1}{Q}$, and for most easy negative pairs, $P_{in}\to 0$. Both two cases will make the contrastive gradient} Eq.~(\ref{eq_grad}) \textit{vanish, damage or disturb the whole learning process.}
	
	Based on the above analysis, we propose Supervised hard positive and negative Pairs Mining~(\textbf{SPM}). It makes supervised contrastive learning more flexible and effective. Concretely, in our SPM model, we introduce two hyper-parameters, $Q_p$ ($Q_p \leq Q$), and $Q_n$ to represent the cardinality for the positive pairs and the negative pairs, respectively. According to the label $y_i$ of $\bz^{(1)}_i$, we reconstruct a new positive queue by mining the top-$Q_p$ hard positives just from Queue-$k$ (based on the \textit{smaller} similarity values): 
	\begin{equation}
	\mQ_{p} = \text{topk}(-\mQ_{y_i} \cdot \bz^{(1)}_i, Q_p). \notag
	\end{equation}
	Similarly,  we reconstruct a new negative queue by mining the top-$Q_n$ hard negatives from all class-balanced queues except Queue-$k$ (based on the \textit{larger} similarity values):
	\begin{equation}
	\mQ_{n} = \text{topk}((\bigcup\limits_{k\neq y_i}^K \mQ_{k} )\cdot \bz^{(1)}_i, Q_n). \notag
	\end{equation}
	Both the hard positives and negatives are more likely to have larger gradients norm and contain more useful information. Thus, the supervised hard positive and negative pairs mining version of the contrastive loss can be designed as:
	\begin{equation}
		\label{eq_spm}
		\begin{split}
			{L}^{\rm{SPM}}_i &= -\frac{w_{y_i}}{Q_p}\sum_{\bz_p \in \mQ_p}\log\frac{\exp{(\bz^{(1)}_i} \bigcdot \bz_p / \tau)}{{\rm{pos}}(\bz^{(1)}_i) + {\rm{neg}}(\bz^{(1)}_i)},  \\
			{\rm{pos}}(\bz^{(1)}_i)&=\sum_{\bz_p \in \mQ_p}\exp{(\bz^{(1)}_i} \bigcdot \bz_p / \tau),  \\
			{\rm{neg}}(\bz^{(1)}_i)&=\sum_{\bz_n \in \mQ_n}\exp{(\bz^{(1)}_i} \bigcdot \bz_n / \tau). 
		\end{split}
	\end{equation}
	\noindent If we set $Q_p=Q$ and $Q_n=(K-1)Q$, the SPM model will degrade to the class-balanced queue model, which means the SPM model is more flexible and the class-balanced queue is just a special case. Thus, we can specify the number of positive pairs and the number of negative pairs, and mine the hard positives and negatives in the contrastive loss to neglect some unfriendly or noisy pairs for better representation learning and generalization.
	
	\setlength{\textfloatsep}{5pt}
	\begin{algorithm}[t]
		\caption{Pseudocode of ResCom in a PyTorch-like style.}
		\label{alg:code}
		\algcomment{\fontsize{8pt}{0em}\selectfont \texttt{mm}: matrix multiplication; \texttt{cat}: concatenation; \texttt{zeros}/\texttt{ones}: a tensor filled with 0/1;  \texttt{topk}: returns the $k$ largest elements of the given input.
		}
		
		\definecolor{codeblue}{rgb}{0.25,0.5,0.5}
		\lstset{
			backgroundcolor=\color{white},
			basicstyle=\fontsize{7pt}{7pt}\ttfamily\selectfont,
			columns=fullflexible,
			breaklines=true,
			captionpos=b,
			commentstyle=\fontsize{7pt}{7pt}\color{codeblue},
			keywordstyle=\fontsize{7pt}{7pt},
		}
\begin{lstlisting}[language=python]
# encoder: encoder network
# FC: a single linear layer classifier (D x K)
# MLP: projection head (including L2 normalization)
# Qp: the number of the hard positive pairs
# Qn: the number of the hard negative pairs
# K: class number, t: temperature
# lambda: loss weight, beta: effective number parameter
# queue: a class-balanced queue of keys (C x KQ)

for x in loader:  # load a minibatch x with N samples
    x1, x2 = aug(x), aug(x)  # randomly augmented

    # Siamese forward
    f1, f2 = encoder(x1), encoder(x2)    		 # features: N x D
    s1, s2 = FC(f1), FC(f2)     			 # classified logits: N x K
    z1, z2 = MLP(f1), MLP(f2)  		 # keys and queries: N x C

    # supervised hard positive and negative pairs mining
    # pick top-Qp smallest pos. logit: N x Qp
    logits_pos = topk(-mm(z1, GetPos(queue, labels)), Qp)
    # pick top-Qn largest neg. pairs: N x Qn
    logits_neg = topk( mm(z1, GetNeg(queue, labels)), Qn)

    # contrastive logits: N x (Qp + Qn)
    logits_cont = cat([logits_pos, logits_neg], dim=1)

    # class-balanced supervised contrastive loss
    mask = cat([ones(N, Qp), zeros(N, Qn)], dim=1)
    l_cont = CBSCLoss(logits_cont / t, beta, mask, labels)

    # Siamese Balanced Softmax
    l_cls = BalSfxLoss(s1, labels) + BalSfxLoss(s2, labels)

    # total loss
    loss = 0.5 * l_cls + lambda * l_cont

    # SGD update: encoder, MLP, and the FC classifier
    loss.backward(), optimizer.step()

    # update the class-balanced queue
    dequeue_enqueue(queue, z2, labels)
\end{lstlisting}
	\end{algorithm}
	
	
	\subsection{Siamese Balanced Softmax}
	
	The two image features $\ff_i^{(1)}, \ff_i^{(2)}$ are sent to a shared classifier to get two classified logits $\bs_i^{(1)}$ and $\bs_i^{(2)}$. Since $\bs_i^{(1)}$ and $\bs_i^{(2)}$ are come from the same input $\bx_i$ and shared the same label $y_i$, we expect to \textit{maximize the mutual information between the two view classified logits}: $I(\bs_i^{(1)}, \bs_i^{(2)})$. However, directly computing the mutual information on imbalanced datasets is not easy. Here we introduce an engineering simplifying assumption to approximately measure it: $I(\bs_i^{(1)}, \bs_i^{(2)})\approx c-H(\bs_i^{(1)},y_i) - H(\bs_i^{(2)},y_i)$, where $H$ represents the Balanced Softmax version of cross-entropy~\cite{balsfx}, and $c$ is a constant. Thus, maximizing the mutual information between the representations of two views is equivalent to minimizing the cross-entropy of them.  Ignoring the constant term $c$,  we get Siamese Balanced Softmax~(\textbf{SiamBS}) to approximate the negative mutual information:
	\begin{equation}
		L_i^{\text{SiamBS}} = -\frac{1}{2}\sum_{v=1}^2\log\left(\frac{N_{y_i}\exp{(\bs_{i, y_i}^{(v)})}}{\sum_{k=1}^{K}N_k\exp{(\bs_{i,k}^{(v)})}}\right),
	\end{equation}
	\noindent where $N_{k}$ is the number of samples in Class-$k$, and $v$ is the view index and can only take values from $\{1, 2\}$.


	\subsection{Overview}
	\vspace{0.5em}
	\noindent
	\textbf{Training.}
	 Fig.~\ref{fig_framework}(c) shows the overview of the proposed ResCom for long-tailed recognition. The framework consists of two Siamese branches for contrastive representation learning and classification learning. 
	We use the supervised hard positive and negative pairs mining version of contrastive loss, Eq.~(\ref{eq_spm}), as the regularization term for representation learning. Thus, the total loss of ResCom can be written as follows:
	\begin{equation}
		{L}^{\rm{ResCom}}_i = L_i^{\text{SiamBS}} + \lambda{L}^{\rm{SPM}}_i(\bz^{(1)}_i, y_i, \bigcup\limits_{k=1}^K \mQ_{k}, Q_p, Q_n),
	\end{equation}
	where $\lambda$ is a loss weight hyper-parameter. Unlike~\cite{supcon, moco, yang2020rethinking}, the ResCom loss allows us to train the model for recognition \textit{in one stage} without linear probing or fine-tuning. Algorithm~\ref{alg:code} provides the pseudo-code of ResCom for detailed reference.
	
	\vspace{0.5em}
	\noindent
	\textbf{Inference.}
	When using ResCom for inference, we only \textit{preserve a single classification branch}. It means that just the encoder and the classifier are preserved, while the MLP is discarded. Therefore, the evaluation of ResCom is very efficient: It is consistent with conventional backbone like the plain ResNet~\cite{resnet}, \textit{without any additional computations}.

	\section{Experiments} 

	\subsection{Datasets and Evaluation Protocol}
	Our experimental setup including the implementation details mainly follows the previous studies~\cite{ldam, mislas, balsfx, paco} for CIFAR-10-LT, CIFAR-100-LT, ImageNet-LT, Places-LT, and iNuturalist 2018. After long-tailed training, we evaluate the models on the corresponding balanced validation datasets, and report the commonly used top-1 accuracy over all classes, denoted as \textbf{All}. We also follow~\cite{liu2019large, lws} to split the
	categories into three subsets and report the average accuracy rates in these three subsets: \textbf{Many}-shot (>100 images), \textbf{Medium}-shot (20-100 images), and \textbf{Few}-shot (<20 images), which are also called the head, medium and tail categories, respectively. 
	
	\vspace{0.5em}
	\noindent
	\textbf{CIFAR-10 \& CIFAR-100.} CIFAR-10 and CIFAR-100 both have 60,000 images --– 50,000 for training and 10,000 for validation with 10 and 100 categories, respectively. For a fair comparison, we use the long-tailed version of CIFAR datasets with the same setting as those used in~\cite{bbn, effnum, balsfx}. They control the degrees of data imbalance with an imbalance factor, $\frac{N_{\text{max}}}{N_{\text{min}}}$. Following~\cite{effnum}, we conduct experiments with imbalance factors 100, 50, and 10. 
	
	\vspace{0.5em}
	\noindent
	\textbf{ImageNet-LT.} 
	ImageNet-LT was proposed by Liu \etal~\cite{liu2019large}. It is a long-tailed version of the large-scale object classification dataset ImageNet~\cite{imagenet}, by sampling a subset following the Pareto distribution with a power value equaling to 6. It contains 115.8K images from 1,000 categories and its imbalance factor is 256, with class cardinality ranging from 5 to 1,280.
	
	\vspace{0.5em}
	\noindent
	\textbf{Places-LT.} 
	Places-LT is similar to ImageNet-LT.  It was also proposed by Liu \etal~\cite{liu2019large}, and is a long-tailed version of the large-scale scene classification dataset Places~\cite{places}. It consists of 184.5K images from 365 categories and its imbalance factor is 996, with class cardinality ranging 5 to 4,980.

	\vspace{0.5em}
	\noindent
	\textbf{iNaturalist 2018.} 
	iNaturalist 2018~\cite{van2018inaturalist} is a species classification dataset, which is on a large scale and suffers from extremely imbalanced label distribution. It is composed of 437.5K images from 8,142 categories and its imbalance factor is 500, with class cardinality ranging from 2 to 1,000. In addition to the extreme imbalance, iNaturalist 2018 also confronts the fine-grained problem~\cite{wei2019piecewise}.

	\subsection{Implementation Details}
	
	As Chen~\etal~\cite{simclr} concluded, contrastive learning usually benefits from longer training times compared with traditional supervised learning with the cross-entropy loss, which is also validated by previous work, \eg, MoCo~\cite{moco}, BYOL~\cite{byol}, SWAV~\cite{caron2020unsupervised}, and SimSiam~\cite{simsiam}. They trained their models in 800 epochs for convergence. Supervised contrastive learning~\cite{supcon} also trains 350 epochs for feature representation learning and another 350 epochs for classification learning.  Thus, we run ResCom with 400 epochs on CIFAR-10-LT, CIFAR-100-LT, ImageNet-LT, and iNaturalist 2018 for better converge of contrastive learning. For Places-LT, we follow previous work~\cite{lws, paco} by loading the pre-trained model from the full ImageNet dataset and fine-tuning it for 30 epochs on Places-LT to \textit{prevent over-fitting}. All models are trained using the SGD optimizer with momentum $\mu = 0.9$. For simplicity, we set the loss weight hyper-parameter $\lambda=0.5$ and the contrastive loss temperature $\tau=0.2$. 
	
	\vspace{0.5em}
	\noindent
	\textbf{CIFAR-10-LT \& CIFAR-100-LT.} 
	We use ResNet-32 as the backbone and strictly follow the setting of \cite{balsfx, paco} for fair comparison. We train ResCom on one GPU with batch size 128. The learning rate decays by a multi-step scheduler from 0.1 to 0.001 at the 320-th epoch and the 360-th epoch.
	
	\vspace{0.5em}
	\noindent
	\textbf{ImageNet-LT.} 
	We used ResNet-50 and ResNeXt-50 as our backbones and strictly follow the setting of \cite{ride, paco} for fair comparison. We train ResCom on 8 GPUs with batch size 256. The learning rate decays by a cosine scheduler from 0.06 to 0 for 400 epochs.

	\vspace{0.5em}
	\noindent
	\textbf{Places-LT.} 
	Following previous setting \cite{lws, zhang2021distribution, lade, mislas, paco}, we choose ResNet-152 as the backbone network, which is pre-trained from the full ImageNet dataset (provided by PyTorch~\cite{paszke2019pytorch}). We fine-tune it for 30 epochs. Similar to ImageNet-LT, the learning rate decays by a cosine scheduler from 0.02 to 0 with batch size 256 on 8 GPUs. 
	
	\vspace{0.5em}
	\noindent
	\textbf{iNaturalist 2018.} 
	Following the previous setting \cite{ride, paco}, we conduct experiments with ResNet-50. Similar to ImageNet-LT, the learning rate decays by a cosine scheduler from 0.08 to 0 with batch size 256 on 8 GPUs for 400 epochs.

	\begin{figure*}[t]
		\centering
		\includegraphics[width=0.99\linewidth]{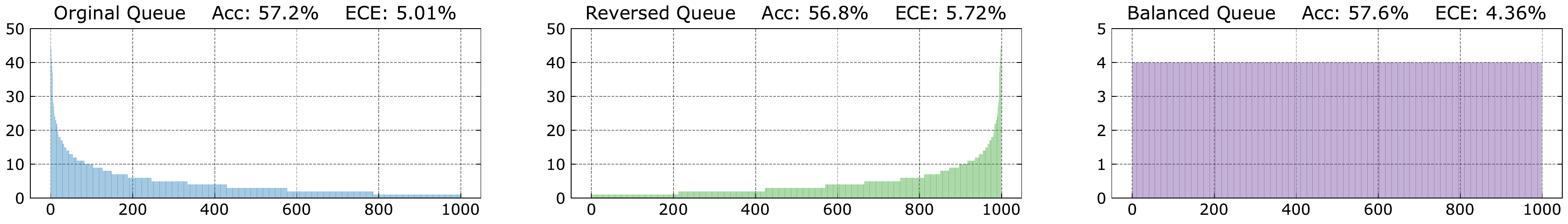}\\
		\caption{Class distribution visualization of different types of queues. The sizes of different queues are close to 4,000. From left to right: original queue, reversed queue, and class-balanced queue. The classes are sorted by descending values of the number of samples per class. Top-1 accuracy and ECE comparisons for different types of queues on \textbf{ImageNet-LT} with \textbf{ResNet-50} are listed at the top.}
		\label{fig:ab_q}
	\end{figure*}

	\begin{figure*}[t]
		\centering
		\includegraphics[width=0.99\linewidth]{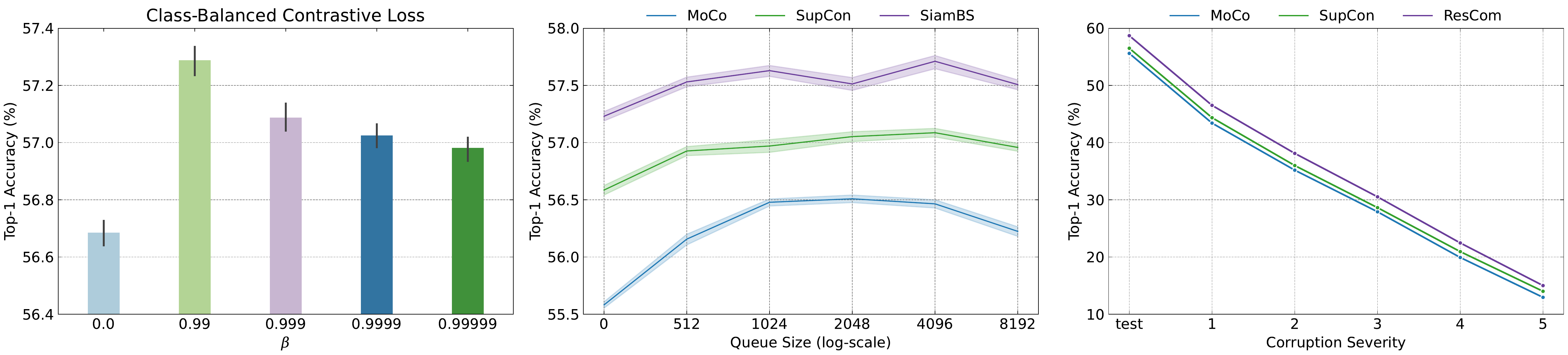}\\
		\caption{Left: ablation experiment of the class-balanced contrastive loss for different $\beta$ on \textbf{ImageNet-LT} with \textbf{ResNet-50}. Center: ablation of three joint contrastive and classification learning mechanisms, SupCon + Balanced Softmax, MoCo + Balanced Softmax, and SupCon + SiamBS on \textbf{ImageNet-LT} with \textbf{ResNet-50}. Right: mean accuracy as a function of corruption severity averaged over all various	corruptions (higher is better) on \textbf{ImageNet-C} with \textbf{ResNet-50}.}
		\label{fig:ab}
	\end{figure*}

	\subsection{Ablation Study}\label{sec_ab}

	\subsubsection{Class-balanced Contrastive Loss}
	 
	In our class-balanced contrastive loss, there is one hyper-parameter $\beta$ in Eq. (\ref{eqn:cbsc}), which controls the weight penalty of classes. It’s worth noting that, $\beta = 0$ corresponds to no reweighting (each class is treated equally), and $\beta \rightarrow 1$ corresponds to reweighting by the inverse class frequency. Here we conduct experiments by varying the $\beta$ from 0.0 to 0.99999 on ImageNet-LT with ResNet-50. We plot the performances upon $\beta$ in Fig.~\ref{fig:ab}~(left) for several possible variants. It shows that the top-1 accuracy can be further improved by \textbf{0.5\%} compared with the conventional supervised contrastive loss ($\beta=0$) when we pick $\beta=0.99$ for the class-balanced contrastive loss. \textit{Consistent improvements} are yielded when picking $\beta$ for other possible values. However, when $\beta$ becomes larger (the class weight distribution becomes sharper), the performances degrade compared  with $\beta=0.99$.

	\subsubsection{Class-balanced Queue}

	In this part, we verify the performances of the class-balanced queue. We define three types of queues here, original queue, reversed queue, and class-balanced queue. There is no constraint on the original queue. Thus, the class distribution of the original queue follows the long-tailed distribution. For the reversed queue, we constrain the class number for each class. The class number of the reversed queue follows the reversed long-tailed class distribution, which means the queue will include more instances for the tail classes than for head classes. As mentioned in Sec.~\ref{bq}, we construct the class-balanced queue with an equal class number for all classes. We draw the class distribution visualization of the above three types of queues in Fig.~\ref{fig:ab_q}. For a fair comparison, the queue sizes for different types are close to 4,000. We train these three variants with the class-balanced contrastive loss on \textbf{ImageNet-LT} with \textbf{ResNet-50}. The detailed performance results are listed at the top of Fig.~\ref{fig:ab_q}. From it, the class-balanced queue can \textit{achieve the best recognition accuracy}. Moreover, we also measure the expected calibration error~(ECE) for these variants to measure the confidence calibration performances on long-tailed recognition~\cite{mislas}. The results also show that the class-balanced queue also \textit{increases the calibration robustness} for long-tailed learning.  Perhaps due to the fact that there are too few positive contrastive pairs, the reversed queue has not achieved very good results, and its performance is even worse than that of the original queue.

	\subsubsection{Supervised Hard Positive and Negative Pairs Mining}
		
	In our SPM model, there are two hyper-parameters $Q_p$ and $Q_n$ to control the mining process of the positive contrastive pairs and negative contrastive pairs. Here we conduct experiments to test the effectiveness of SPM. We train ResCom with different $Q$, $Q_p$, and $Q_n$ on \textbf{ImageNet-LT} with \textbf{ResNeXt-50}. We list the top-1 accuracy results upon $Q$, $Q_p$, and $Q_n$ in Table~\ref{ab_spm} for some possible variants. As we mentioned in the Sec~\ref{sec:spm}, if $Q_p=Q$ and  $Q_n=K(Q-1)$, the SPM model will degrade to the plain class-balanced queue model. According to the results (Row1, Row-3 and Row-4, Row-7), the SPM model can further improve the recognition performance by about $0.8\%$ compared with the plain class-balanced queue model. We also observe that smaller $Q_p$ and $Q_n$ tend to achieve better results: We learn that, not all positives and negatives offer significant contributions to the final performances and most positives and negatives do not help a lot towards the whole learning process. Based on mining an appropriate part (usually small) of hard positives and negatives, the model can learn better representative features and further improve the performance.
	
		\begin{figure*}[t]
		\begin{center}
			\includegraphics[width=0.983\linewidth]{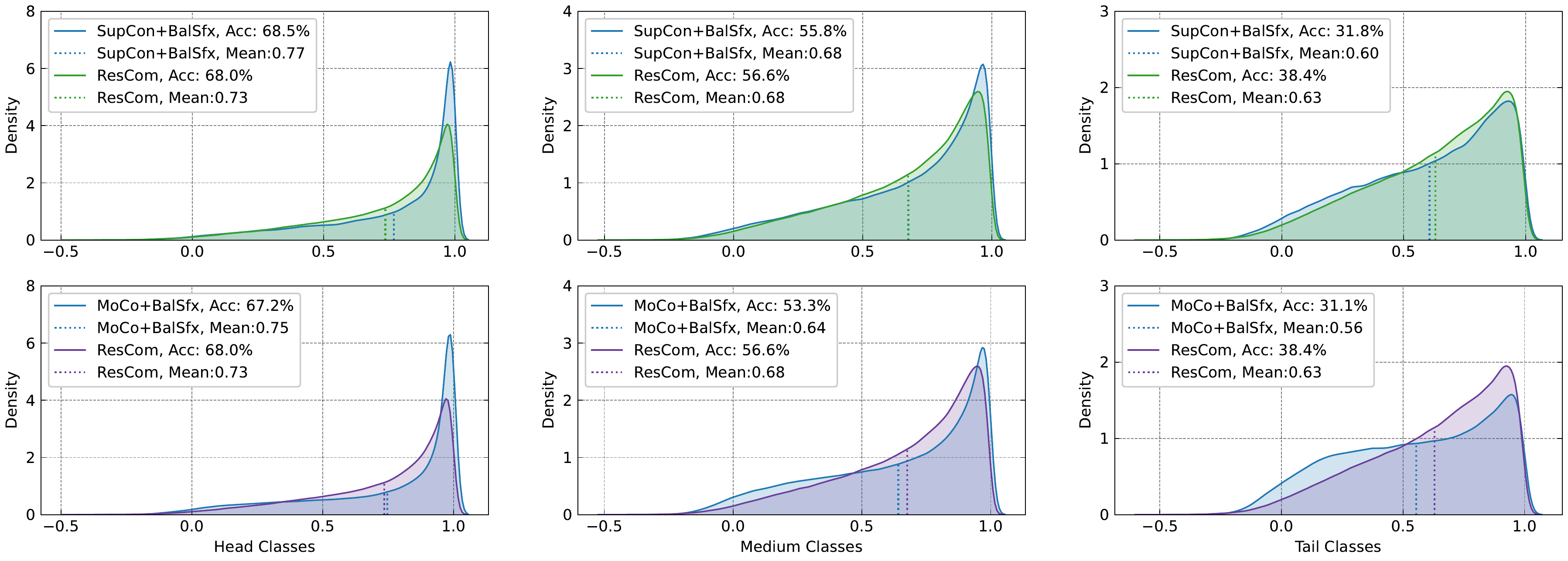}
		\end{center}
		\caption{Similarity distribution of the positives on the \textbf{ImageNet-LT validation} dataset with \textbf{ResNet-50}. We compare the results of three methods, SupCon + Balanced Softmax, MoCo + Balanced Softmax, and ResCom. From left to right: head classes, medium classes, and tail classes. ResCom maintains the accuracy of the head classes while improving the accuracy of the medium and tail classes (larger mean, more positive contrastive pairs get higher similarities, \textit{best viewed in color}). }
		\label{fig:pos_sim}
	\end{figure*}
	
	\begin{figure*}[t]
		\centering
		\includegraphics[width=0.983\linewidth]{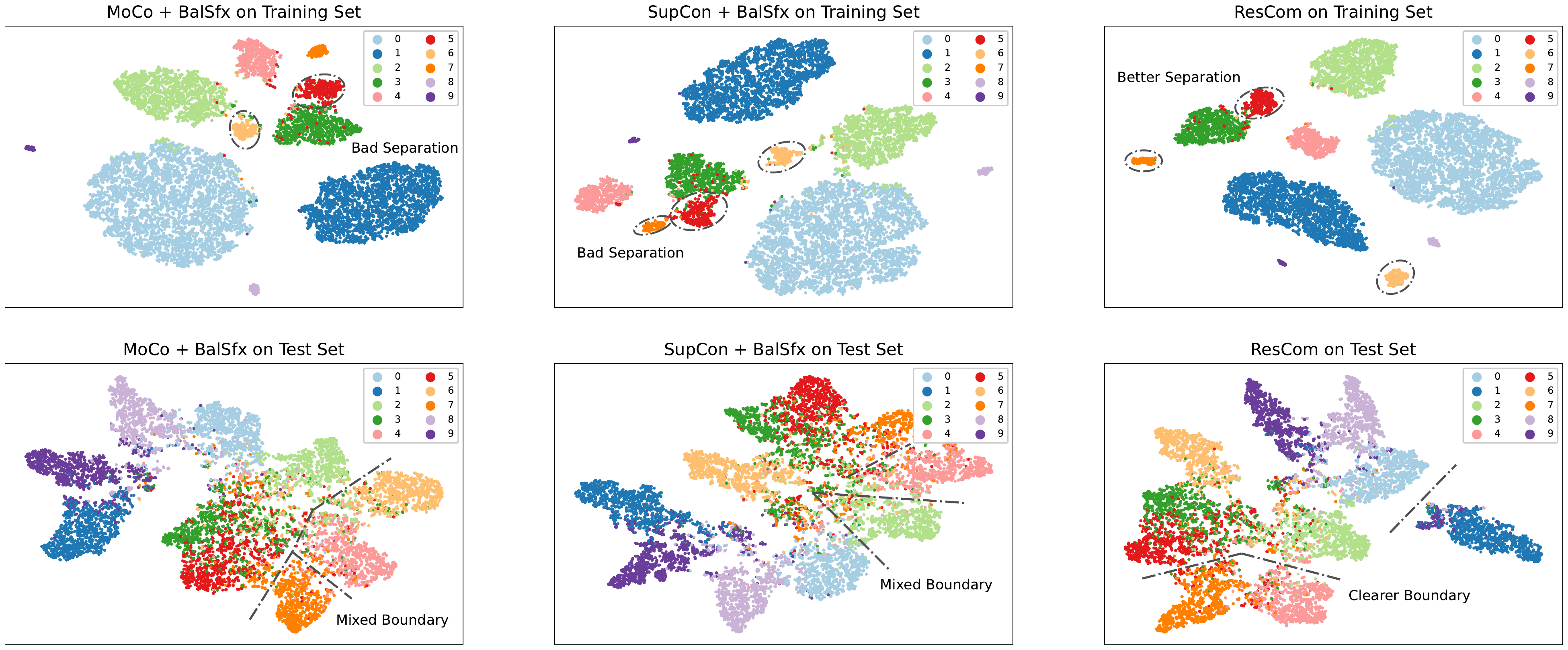}
		\caption{t-SNE visualization of training \& test set on \textbf{CIFAR-10-LT}, \textbf{imbalanced factor 100}, with \textbf{ResNet-32}. Comparing with  MoCo and SupCon, ResCom helps mitigate the medium and the tail classes leakage during testing, which results in better learned boundaries and representations. Numbers at the top right of the figures are class indexes (\textit{best viewed in color}).}
		\label{fig:ablation_study_feature_visualization}
		
	\end{figure*}
	
	\subsubsection{Siamese Balanced Softmax}
	
	In the final framework of our ResCom, we involve Siamese Balanced Softmax~(SiamBS) in our total loss. To verify the power of the SiamBS part, we build two baseline models here. In the first model, we adopt the Balanced Softmax loss for single view (on $\bs^{(1)}$) classification learning and SupCon for additional contrastive regularization. In the second model,  we adopt the Balanced Softmax loss for single view classification learning and MoCo style contrastive learning for the regularization. We build our model with SiamBS for double views classification learning and SupCon for the contrastive regularization. We draw the results in Fig.~\ref{fig:ab}~(right).  From it, our SiamBS yields consistent improvements \textbf{0.6\%} on ImageNet-LT with the ResNet-50 backbone for all possible queue sizes over the SupCon and MoCo based models. The above results firmly manifest the effectiveness of SiamBS.

	\subsubsection{Similarity Distribution and Embedding Visualizations}
	
			\begin{table*}[t]
		\begin{minipage}[t]{0.31\textwidth}
			\caption{Top-1 accuracy (\%) and inference time (ms) results on \textbf{ImageNet-LT} with \textbf{ResNet-50}.} 
			\begin{tabular}{y{56}y{42}y{25}}
				\toprule
				Method & {Infr. time}  & {Top-1}\\
				\midrule
				RIDE(2 experts)  &12.0\cgaphll{+}{44\%~~}  & 54.4\\
				RIDE(3 experts)  &15.3\cgaphll{+}{64\%~~}  & 54.9\\
				RIDE(4 experts)  &18.5\cgaphll{+}{122\%}   & 55.4\\
				\cellcolor{Gray}\textbf{ResCom} & \cellcolor{Gray}\textbf{8.3} & 
				\cellcolor{Gray}\textbf{58.7} \\
				\bottomrule
			\end{tabular}
			\label{ride1}
		\end{minipage}
		\hfill
		\begin{minipage}[t]{0.31\textwidth}
			\caption{Top-1 accuracy (\%) and inference time (ms) results on \textbf{ImageNet-LT} with \textbf{ResNeXt-50}.} 
			\begin{tabular}{y{56}y{42}y{25}}
				\toprule
				Method & {Infr. time}  & {Top-1}\\
				\midrule
				RIDE(2 experts)  &19.0\cgaphll{+}{46\%~~} & 55.9\\
				RIDE(3 experts)  &26.0\cgaphll{+}{100\%}  & 56.4\\
				RIDE(4 experts)  &33.2\cgaphll{+}{155\%}  & 56.8\\
				\cellcolor{Gray}\textbf{ResCom} & \cellcolor{Gray}\textbf{13.0} & 
				\cellcolor{Gray}\textbf{59.2} \\
				\bottomrule
			\end{tabular}
			\label{ride2}
		\end{minipage}
		\hfill
		\begin{minipage}[t]{0.31\textwidth}
			\centering
			\caption{Robustness as measured by Mean Corruption Error (mCE) and Normalized mCE over \textbf{ImageNet-C} with \textbf{ResNet-50} (lower is better).} 
			\vspace{-2.4pt}
			\begin{tabular}{y{50}x{21}x{21}x{21}}
				\toprule
				{Method} & {Top-1}   & mCE & NmCE \\
				
				\midrule
				MoCo+BalSfx  & 55.7  & 71.4 & 91.4\\
				SupCon+BalSfx  &56.5 & 71.1 & 91.2\\
				\cellcolor{Gray}\textbf{ResCom}  &\cellcolor{Gray} \textbf{58.7}  & \cellcolor{Gray} \textbf{69.4} & \cellcolor{Gray}\textbf{89.2} \\
				\bottomrule
			\end{tabular}
			\label{mce}
		\end{minipage}
	\end{table*}
	
	\begin{table*}[h]
		\centering
		\renewcommand\tabcolsep{11.0pt}
		\renewcommand\arraystretch{1.1}
		\caption{Orthogonality to multi-stage methods. The performances can be further improved by adding two-stage methods.}
		\begin{tabular}{y{70}|y{20}y{20}y{20}y{40}|y{20}y{20}y{20}y{40}}
			\toprule
			\multirow{2.5}{*}{Method} &  \multicolumn{4}{c|}{ResNet-50} & \multicolumn{4}{c}{ResNeXt-50}\\
			\cmidrule(lr){2-9}
			& Many &Med.  &Few  &\textbf{All}  & Many &Med.  &Few  &\textbf{All} \\
			\midrule
			
			ResCom               &68.0 &56.6 &38.6 &58.7 & 69.6 & 56.6 & 39.1 & 59.2\\
			ResCom + cRT     &68.6 &55.7 &36.9 &58.4 & 70.9 & 55.3 & 33.5 & 58.4\\
			ResCom + LWS     &68.3 &55.8 &38.8 &58.8 &70.1 & 55.7 & 37.8 & 58.8\\
			ResCom + LAS     &67.8 &  57.4& 40.6& \textbf{59.2}\cgaphl{+}{0.5} &68.7 & 58.1 & 39.5 &\textbf{59.5}\cgaphl{+}{0.3}\\
			\bottomrule
		\end{tabular}
		\label{orth}
	\end{table*}

	\begin{table*}[t]
		\begin{minipage}{0.32\textwidth}
			\centering
			\renewcommand\arraystretch{1.1}
			\caption{Ablation study of the SPM model for different $Q$, $Q_p$, and $Q_n$.}
			\begin{tabular}{y{30}x{25}y{25}y{40}}
				\toprule
				$K \times Q$ & $Q_p$   & $Q_n$ & Top-1 Acc.\\
				\midrule
				\cellcolor{Gray} $1000 \times 4$ & \cellcolor{Gray} 1 & \cellcolor{Gray} 500 & \cellcolor{Gray} \textbf{59.2}\cgaphl{+}{0.9} \\
				$1000 \times 4$ & 2 & 2000 & 58.6 \\
				\cellcolor{Gray} $1000 \times 4$ & \cellcolor{Gray} 4 & \cellcolor{Gray} 3996 & \cellcolor{Gray} 58.3 \\
				\midrule
				\cellcolor{Gray} $1000 \times 8$ & \cellcolor{Gray} 2 & \cellcolor{Gray} 500 & \cellcolor{Gray} \textbf{59.0}\cgaphl{+}{0.8} \\
				$1000 \times 8$ & 2 & 2000 & 58.7 \\
				$1000 \times 8$ & 4 & 2000 & 58.4 \\
				\cellcolor{Gray} $1000 \times 8$ & \cellcolor{Gray} 8 & \cellcolor{Gray} 7992 & \cellcolor{Gray} 58.2 \\
				\bottomrule
			\end{tabular}
		\label{ab_spm}
		\end{minipage}
		\hfill
		\begin{minipage}{0.62\textwidth}
			\centering
			\renewcommand\arraystretch{1.03}
			\caption{Ablation study for all proposed
				modules on \textbf{ImageNet-LT} with \textbf{ResNet-50}. CBL: Class-balanced SupCon loss. CBQ: Class-balanced queue. }
			\begin{tabular}{x{30}x{30}|x{16}x{16}x{16}|x{24}x{24}|x{59}}
				\toprule
				\multicolumn{2}{c|}{Framework}	  & \multicolumn{3}{c|}{Contrastive Module}	  & \multicolumn{2}{c|}{Classified Module}	  & {\multirow{2.5}{*}{{Top-1 Acc. }}} \\
				\cmidrule(lr){1-7}
				MoCo & SupCon & CBL & CBQ & SPM & BalSfx  & SiamBS &  \\
				\midrule
				
				{\scriptsize\ding{52}} & {\scriptsize \ding{56}} &  {\scriptsize \ding{56}}& {\scriptsize \ding{56}} & {\scriptsize \ding{56}}  & {\scriptsize\ding{52}} & {\scriptsize \ding{56}} & 
				55.7 \\
				
				{\scriptsize \ding{56}} & {\scriptsize\ding{52}} & {\scriptsize \ding{56}} &  {\scriptsize \ding{56}}& {\scriptsize \ding{56}} & {\scriptsize\ding{52}}  &  {\scriptsize \ding{56}} & 
				56.5 \\
				\midrule
				{\scriptsize \ding{56}} & {\scriptsize\ding{52}} & {\scriptsize\ding{52}} &  {\scriptsize \ding{56}}& {\scriptsize \ding{56}} & {\scriptsize\ding{52}}  &  {\scriptsize \ding{56}} & 
				57.2 \\
				
				{\scriptsize \ding{56}} & {\scriptsize\ding{52}} & {\scriptsize\ding{52}} & {\scriptsize\ding{52}} & {\scriptsize \ding{56}} & {\scriptsize\ding{52}}  &  {\scriptsize \ding{56}} & 
				57.6 \\
				
				{\scriptsize \ding{56}} & {\scriptsize\ding{52}} & {\scriptsize\ding{52}} & {\scriptsize\ding{52}} & {\scriptsize\ding{52}} & {\scriptsize\ding{52}}  &  {\scriptsize \ding{56}} & 
				58.1 \\

				\multicolumn{1}{>{\columncolor{Gray}[6pt][282pt]}c}{\scriptsize \ding{56}} & {\scriptsize \ding{52}} & {\scriptsize \ding{52}} & {\scriptsize \ding{52}} & {\scriptsize \ding{52}} & {\scriptsize \ding{56}} & {\scriptsize \ding{52}} &
				\hspace{20pt} \textbf{58.7}\cgaphl{+}{2.2} \\
				\bottomrule
			\end{tabular}
			\label{ab_all}
		\end{minipage}
	\end{table*}

	The above experiments show the effectiveness of the proposed three components, \ie, class-balanced contrastive loss, class-balanced queue, and Siamese Balanced Softmax in ResCom. In this part, we combine these three components together to visualize the contrastive similarity distribution of ResCom. We set two baseline models: joint training with contrastive loss and Balanced Softmax loss under the MoCo~\cite{moco}  framework and the SupCon~\cite{supcon} framework. The MoCo framework can achieve 55.7\% top-1 accuracy, and the SupCon framework can achieve 56.5\% top-1 accuracy on ImageNet-LT. Under the same setting, ResCom can achieve surprising \textbf{58.7\%} top-1 accuracy, which is more than \textbf{2\%} higher than the results of two baseline models. We draw the similarity distribution of the positive contrastive pairs on the ImageNet-LT validation dataset in Fig.~\ref{fig:pos_sim}. ResCom maintains the accuracy of the head classes and improves the performances of the medium and tail classes at the same time. Using ResCom, the similarity distributions of the head classes has a smaller mean value, and the similarity distributions of the medium classes and the tailed classes have larger mean values, and \textit{more positive pairs close to 1}), which means ResCom \textit{rebalances the similarity distributions} and is more friendly to tailed classes.

	Further, we look at the t-SNE projection of learned representations for the above two baselines and ResCom. For each method, the projection is performed over both training and test data from \textbf{CIFAR-10-LT}, \textbf{imbalanced factor 100}. Thus providing the same decision boundary for better
	visualization. Fig.~\ref{fig:ablation_study_feature_visualization} shows that the decision boundaries and class separations of two baselines can be greatly altered by the
	head classes, which results in the large leakage of tail classes during (balanced) inference. In contrast, ResCom sustains better separation with less leakage, especially between adjacent head and tail class.
	
	\subsubsection{Inference Time Analysis}
	
		One more thing we would like to mention is the RIDE-based~\cite{ride} ensemble (or mixture of experts) methods have higher inference latency than standard network based methods like ResCom. The inference time comparison results are listed in Table~\ref{ride1} and \ref{ride2}. From it, RIDE~\cite{ride} (or DiVE~\cite{he2021distilling}) with four experts takes \textbf{18.5ms} for inference with a batch of 64 images on the Nvidia GeForce 2080Ti GPU.  While ResCom is a single model and just takes \textbf{8.3ms} under the same situation for \textbf{ResNet-50}. For \textbf{ResNeXt-50}, RIDE four experts takes \textbf{33.2ms},  while ResCom just takes \textbf{13.0ms} under the same situation. Moreover, ResCom can achieve much better recognition results over RIDE-based methods. The same conclusions can  be also obtained on other datasets like iNaturalist 2018.

	\subsubsection{Robustness to Image Corruptions}
	
	Deep neural networks lack robustness to out of distribution data or natural corruptions such as noise, blur and JPEG compression. The benchmark ImageNet-C dataset~~\cite{hendrycks2019benchmarking} is used to measure trained model performance on such corruptions.  Following \cite{supcon, hendrycks2019benchmarking}, we compare the ResCom models to MoCo, SupCon and Balanced Softmax~\cite{balsfx} using the Mean Corruption Error (mCE) and Normalized (by AlexNet~\cite{imagenet}) Mean Corruption Error~(NmCE) metrics. Both metrics measure average degradation in performance compared to ImageNet test set, averaged over all possible corruptions and severity levels. The quantitative results are listed in Table~\ref{mce}. The ResCom model has better top-1 accuracy and lower mCE and NmCE values across different corruptions, showing increased robustness. In Fig.~\ref{fig:ab} (right), we also find that ResCom models demonstrate lesser degradation in accuracy with increasing corruption severity. 
%
			\begin{table*}[t]
		\begin{minipage}[t]{0.48\textwidth}
			\centering
			\renewcommand\tabcolsep{8.0pt}
			\renewcommand\arraystretch{1.1}
			\caption{Top-1 accuracy (\%) comparison on \textbf{CIFAR-10-LT} with \textbf{ResNet-32} for different
				imbalance factors.}
			\vspace{-2.5pt}
			\begin{tabular}{y{65}y{37}y{37}y{37}}
				\specialrule{0.3pt}{0.1pt}{0pt}
				\toprule
				Imbalanced Factor & 100 &50  &10 \\
				\midrule
				LDAM             & 77.1 & 81.1 & 88.2 \\
				ELF+LDAM         & 78.1 & 82.4 & 88.0 \\
				RSG              & 79.5 & 82.8 & -    \\
				BBN              & 79.9 & 82.2 & 88.4 \\
				ResLT            & 80.5 & 83.5 & 89.1 \\
				Causal Norm      & 80.6 & 83.6 & 88.5 \\
				Hybrid-PSC       & 81.4 & 85.4 & 91.1 \\
				MiSLAS           & 82.1 & 85.7 & 90.0 \\
				Balanced Softmax & 81.5 & 84.9 & 91.3 \\
				\cellcolor{Gray}\textbf{ResCom} & \cellcolor{Gray}\textbf{84.9}\cgaphl{+}{2.8} &
				\cellcolor{Gray}\textbf{88.0}\cgaphl{+}{2.3}  &
				\cellcolor{Gray}\textbf{92.0}\cgaphl{+}{0.7} \\
				\bottomrule
			\end{tabular}
			\label{tab:cifar10}
		\end{minipage}
		\hfill
		\begin{minipage}[t]{0.48\textwidth}
			\centering
			\renewcommand\tabcolsep{8.0pt}
			\renewcommand\arraystretch{1.1}
			\caption{Top-1 accuracy (\%) comparison on \textbf{CIFAR-100-LT} with \textbf{ResNet-32} for different
				imbalance factors.}
			\vspace{-2.5pt}
			\begin{tabular}{y{65}y{37}y{37}y{37}}
				\specialrule{0.3pt}{0.1pt}{0pt}
				\toprule
				Imbalanced Factor & 100 &50  &10 \\
				\midrule
				BBN              & 42.6 & 47.0 & 59.1 \\
				Causal Norm      & 44.1 & 50.3 & 59.6 \\
				ResLT            & 45.3 & 50.0 & 60.8 \\
				LADE             & 45.4 & 50.5 & 61.7 \\
				Hybrid-PSC       & 45.0 & 49.0 & 62.4 \\
				RIDE({3 experts}) & 48.0 & 51.7 & 61.8 \\
				MiSLAS           & 47.0 & 52.3 & 63.2 \\
				Balanced Softmax & 50.8 & 54.2 & 63.0 \\
				PaCo             & 52.0 & 56.0 & 64.2 \\
				\cellcolor{Gray}\textbf{ResCom} & \cellcolor{Gray}\textbf{53.8}\cgaphl{+}{1.8} &
				\cellcolor{Gray}\textbf{58.0}\cgaphl{+}{2.0}  &
				\cellcolor{Gray}\textbf{66.1}\cgaphl{+}{1.9} \\
				\bottomrule
			\end{tabular}
			\label{tab:cifar100}
		\end{minipage}
	\vspace{-2.5pt}
	\end{table*}

	\begin{table*}[t]
		\begin{minipage}[t]{0.48\textwidth}
			\centering
			\renewcommand\tabcolsep{8.0pt}
			\renewcommand\arraystretch{1.1}
			\caption{Top-1 accuracy (\%) comparison with state-of-the-art methods on \textbf{ImageNet-LT} with \textbf{ResNet-50}.}
			\vspace{-2.5pt}
			\begin{tabular}{y{56}y{22}y{22}y{22}y{37}}
				\toprule
				Method & Many &Med.  &Few  &\textbf{All} \\
				\midrule
				
				cRT                  & 62.5 & 47.4 & 29.5 & 50.3 \\
				LWS                  & 61.8 & 48.6 & 33.5 & 51.2 \\
				ELF                  & 64.3 & 47.9 & 31.4 & 52.0 \\
				MiSLAS               & 61.7 & 51.3 & 35.8 & 52.7 \\
				DisAlign             & 61.3 & 52.2 & 31.4 & 52.9 \\
				DRO-LT               & 64.0 & 49.8 & 33.1 & 53.7 \\
				RIDE({4 experts})     & 66.2 & 52.3 & 36.5 & 55.4 \\
				PaCo                 & 65.0 & 55.7 & 38.2 & 57.0 \\
				
				\cellcolor{Gray}\textbf{ResCom} & \cellcolor{Gray}\textbf{68.0} &
				\cellcolor{Gray}\textbf{56.6} &
				\cellcolor{Gray}\textbf{{38.6}} &
				\cellcolor{Gray}\textbf{58.7}\cgaphl{+}{1.7} \\
				\bottomrule
			\end{tabular}
			\label{tab:imgnet_r50}
		\end{minipage}
		\hfill
		\begin{minipage}[t]{0.48\textwidth}
			\centering
			\renewcommand\tabcolsep{8.0pt}
			\renewcommand\arraystretch{1.1}
			\caption{Top-1 accuracy (\%) comparison with state-of-the-art methods on \textbf{ImageNet-LT} with \textbf{ResNeXt-50}.}
			\vspace{-2.5pt}
			\begin{tabular}{y{63}y{20.5}y{20.5}y{20.5}y{37}}
				\toprule
				Method & Many &Med.  &Few  &\textbf{All} \\
				\midrule
				
				LWS      &60.5 &47.2 &31.2 &50.1 \\
				Causal Norm & 65.2& 47.7& 29.8 &52.0\\
				Balanced Softmax &63.6 &48.4 &32.9 &52.1\\
				LADE    &65.1 &48.9 &33.4 &53.0 \\
				DiVE  & 64.1 &  50.5 & 31.5&  53.1 \\
				SSD &66.8&53.1&35.4&56.0\\
				RIDE{(4 experts)}      &68.2 & 53.8 & 36.0 & 56.8\\
				PaCo                  & 67.5 & \textbf{56.9} & 36.7 & 58.2 \\
				
				\cellcolor{Gray}\textbf{ResCom} & \cellcolor{Gray}\textbf{69.5} &
				\cellcolor{Gray}56.7 &
				\cellcolor{Gray}\textbf{39.1}&
				\cellcolor{Gray}\textbf{59.2}\cgaphl{+}{1.0} \\
				\bottomrule
			\end{tabular}
			\label{tab:imgnet_x50}
		\end{minipage}
	\vspace{-2.5pt}
	\end{table*}

			\begin{table*}[t]
		\begin{minipage}[t]{0.48\textwidth}
			\centering
			\renewcommand\tabcolsep{8.0pt}
			\renewcommand\arraystretch{1.1}
			\caption{Top-1 accuracy (\%) comparison on \textbf{Places-LT}, starting from an ImageNet pre-trained \textbf{ResNet-152}.}
			\vspace{-2.5pt}
			\begin{tabular}{y{63}y{20.5}y{20.5}y{20.5}y{37}}
				\specialrule{0.3pt}{0.1pt}{0pt}
				\toprule
				Method & Many &Med.  &Few  &\textbf{All} \\
				\midrule
				LWS                  & 40.6 & 39.1 & 28.6 & 37.6 \\
				$\tau$-norm          & 37.8 & 40.7 & 31.8 & 37.9 \\
				Balanced Softmax     & 42.0 & 39.3 & 30.5 & 38.6 \\
				LADE                 & 42.8 & 39.0 & 31.2 & 38.8 \\
				RSG                  & 41.9 & 41.4 & 32.0 & 39.3 \\
				DisAlign             & 40.4 & 42.4 & 30.1 & 39.3 \\
				GistNet              & 42.5 & 40.8 & 32.1 & 39.6 \\
				ResLT                & 39.8 & 43.6 & 31.4 & 39.8 \\
				MiSLAS               & 39.6 & 43.3 & \textbf{36.1} & 40.4 \\
				PaCo                 & 37.5 & \textbf{47.2} & 33.9 & 41.2 \\
				\cellcolor{Gray}\textbf{ResCom} & \cellcolor{Gray}\textbf{43.0} &
				\cellcolor{Gray}43.4 &
				\cellcolor{Gray}35.3 &
				\cellcolor{Gray}\textbf{41.7}\cgaphl{+}{0.5} \\
				\bottomrule
			\end{tabular}
			\label{tab:places}
		\end{minipage}
		\hfill
		\begin{minipage}[t]{0.48\textwidth}
			\centering
			\renewcommand\tabcolsep{8.0pt}
			\renewcommand\arraystretch{1.1}
			\caption{Top-1 accuracy (\%) comparison with state-of-the-art methods on \textbf{iNaturalist 2018} with \textbf{ResNet-50}.}
			\vspace{-2.5pt}
			\begin{tabular}{y{56}y{22.5}y{22.5}y{22.5}y{37}}
				\specialrule{0.3pt}{0.1pt}{0pt}
				\toprule
				Method & Many &Med.  &Few  &\textbf{All} \\
				\midrule
				ELF                  & 72.7 & 70.4 & 68.3 & 69.8 \\
				LADE                 & -    & -    & -    & 70.0 \\
				RSG				     & -    & -    & -    & 70.3 \\
				DisAlign             & -    & -   & -    & 70.6 \\
				GistNet              & -    & -    & -    & 70.8 \\
				SSD                  & -    & -   & -    & 71.5 \\
				MiSLAS               & 70.4 & 72.4 & 73.2 & 71.6 \\
				RIDE({4 experts})     & 70.9 & 72.4 & 73.1 & 72.6 \\
				PaCo                 & 70.3 & 73.2 & 73.6 & 73.2 \\
				DiVE({4 experts})  	 & -    & -   & -    & 73.4 \\
				\cellcolor{Gray}\textbf{ResCom} & \cellcolor{Gray}\textbf{71.6} &
				\cellcolor{Gray}\textbf{75.0}  &
				\cellcolor{Gray}\textbf{75.6} &
				\cellcolor{Gray}\textbf{75.2}\cgaphl{+}{1.8} \\
				\bottomrule
			\end{tabular}
			\label{tab:ina}
		\end{minipage}
		\vspace{-2.5pt}
	\end{table*}

	\subsubsection{Orthogonality to Multi-stage Methods}
	 Here we explore whether ResCom can be orthogonally applied on existing imbalanced learning methods like multi-stage methods. We mainly add two-stage methods such as cRT~\cite{lws}, LWS~\cite{lws} and LAS~\cite{mislas} to post-processing the ResCom model. We verify the orthogonality on \textbf{ImageNet-LT} with different backbones. The results are listed in Table~\ref{orth}. Some powerful two-stage methods~(LWS and LAS) can further improve the classification performances by \textbf{0.3-0.5\%} with both ResNet-50 and ResNeXt-50. The above experiments verify the orthogonality of our ResCom to multi-stage methods. We leave exploring the orthogonality to other kind of imbalanced methods as our future work.

	\subsubsection{Summary}
	Overall, Table~\ref{ab_all} shows the ablation investigation on the effects of class-balanced supervised contrastive loss~(CBL), class-balanced queue~(CBQ), Supervised Hard Positive and Negative Pairs Mining~(SPM), and Siamese Balanced Softmax~(SiamBS). Comparing with the two baseline models, all proposed modules are added one by one and can consistently improve the top-1 accuracy on \textbf{ImageNet-LT} with \textbf{ResNet-50}. The final ResCom model involves all proposed components can improve the top-1 accuaracy by about \textbf{2\%}.  They firmly manifest the effectiveness of the proposed components in ResCom for imbalanced and long-tailed recognition. 
	
	\subsection{Comparison with State-of-the-arts}
	
	To verify the effectiveness, we compare the proposed method ResCom against four groups of state-of-the-art methods on common-used long-tailed datasets: 
	
	\begin{itemize}
		\item \textbf{Logits modification methods.} For  logits modification methods, we mainly compared with latest LDAM~\cite{ldam}, Causal Norm~\cite{tang2020long}, Balanced Softmax~\cite{balsfx}, and LADE~\cite{lade}. 
		
		\item \textbf{Architecture modification methods.} We mainly compare ResCom with recent architecture modification methods like BBN~\cite{bbn}, ELF~\cite{elf}, ResLT~\cite{cui2021reslt}, GistNet~\cite{liu2021gistnet}, RIDE~\cite{ride}, and DiVE~\cite{he2021distilling}. 
		
		\item \textbf{Multi-stage methods.} We mainly compare ResCom with recent multi-stage methods, such as Decouple models~(cRT, LWS)~\cite{lws}, DisAlign~\cite{zhang2021distribution}, MiSLAS~\cite{mislas}, SSD~\cite{li2021self}.  
		
		\item \textbf{Contrastive learning methods.} We mainly compare ResCom with recent contrastive learning methods, \eg, Hybrid-PSC~\cite{wang2021contrastive}, RSG~\cite{wang2021rsg}, DRO-LT~\cite{samuel2021distributional}, and PaCo~\cite{paco}. 
	\end{itemize}

	\subsubsection{Comparison on CIFAR-LT}
	
	The experimental results on CIFAR-10-LT, CIFAR-100-LT are listed in Table~\ref{tab:cifar10} and \ref{tab:cifar100}, repectively. We mainly compare with the SOTA methods Balanced Softmax~\cite{balsfx} and PaCo~\cite{paco} under the same training setting where Cutout~\cite{devries2017improved} and AutoAugment~\cite{cubuk2018autoaugment} are used in training. As shown in Table~\ref{tab:cifar10} and \ref{tab:cifar100}, ResCom consistently outperforms them across all common-used imbalance factors under such a strong augmentation setting. Concretely, ResCom surpasses the previous best by \textbf{2.8\%}, \textbf{2.3\%} and \textbf{0.7\%} on CIFAR-10-LT, \textbf{1.8\%}, \textbf{2.0\%} and \textbf{1.9\%} on CIFAR-100-LT, under imbalance factor 100, 50, and 10, respectively. It testifies the powerful effectiveness of ResCom.

	\subsubsection{Comparison on ImageNet-LT}
	
	Table~\ref{tab:imgnet_r50} and \ref{tab:imgnet_x50} show extensive experimental results for comparison with recent state-of-the-art methods with the ResNet-50 and ResNeXt-50 backbone on ImageNet-LT. ResCom also surpasses all previous methods by large margins on different backbones: \textbf{1.7\%} for ResNet-50, and \textbf{1.0\%} for ResNeXt-50.

	\subsubsection{Comparison on Places-LT}
	
	The experimental results on Places-LT are summarized in Table~\ref{tab:places}. Our ResCom is flexible to reload the existing pre-trained models because the main network architecture is the same as those of \cite{liu2019large,lws}. However, due to the architecture change of RIDE~\cite{ride} and its variants, it is not applicable to load the publicly pre-trained models.  Under a fair training setting by finely tuning 30 epochs without additional augmentation, ResCom also outperforms all previous methods.

	\subsubsection{Comparison on iNaturalist 2018}		
	
	Table~\ref{tab:ina} lists experimental results on iNaturalist 2018. Under a fair training setting, ResCom consistently surpasses recent state-of-the-art methods \textit{on all accuracy measurements} (many, medium, few, and all). The \textit{single} ResCom model improves top-1 accuracy by a large margin \textbf{1.8\%}.

	\section{Conclusion}
	In this paper, we observe and mathematically analyze that SupCon suffers a \textit{dual class-imbalance} problem at both the \textit{original batch-level} and the \textit{Siamese batch-level}, which is more difficult than it in long-tailed classification learning. We present Rebalanced Siamese Contrastive Mining (ResCom) to address it: we introduce a class-balanced supervised contrastive loss for the original batch-level, a class-balanced queue for the Siamese batch-level.
	Moreover, we noted that easy positives and negatives make the contrastive gradient vanish and may disturb the representation learning. We propose supervised hard positive and negative pairs mining, and Siamese Balanced Softmax to dig for more useful information. Based on the experiments, all proposed modules have been verified to be helpful for long-tailed recognition and \textit{improve the robustness and generalization}. ResCom \textit{effectively rebalances the similarity distributions} and \textit{improves the performances of both the medium and tail classes}, which is consistent with our original goal. Additionally, ResCom \textit{surpasses} the previous competitors by \textit{large margins} on multiple popular long-tailed benchmark datasets.

	\ifCLASSOPTIONcaptionsoff
	\newpage
	\fi
	
	{\small
		\bibliographystyle{IEEEtran}
		\bibliography{egbib}
	}

\end{document}